\documentclass[10pt,twocolumn,letterpaper]{article}

\usepackage{cvpr}
\usepackage{times}
\usepackage{epsfig}
\usepackage{graphicx}
\usepackage{amsmath}
\usepackage{amssymb}
\usepackage[ruled,lined,linesnumbered]{algorithm2e}
\usepackage{multirow}

\usepackage[pagebackref=true,breaklinks=true,letterpaper=true,colorlinks,bookmarks=false]{hyperref}

\cvprfinalcopy 

\def\cvprPaperID{1689} 

\ifcvprfinal\fi
\begin{document}

\title{Highly Efficient Forward and Backward Propagation of Convolutional Neural Networks for Pixelwise Classification}

\author{Hongsheng Li$^{1,2}$, Rui Zhao$^1$, and Xiaogang Wang$^1$\\
$^1$Multimedia Laboratory, The Chinese University of Hong Kong\\
$^2$School of Electronic Engineering, University of Electronic Science and Technology of China\\
{\tt\small lihongsheng@gmail.com, \{rzhao, xgwang\}@ee.cuhk.edu.hk}
}

\maketitle

\begin{abstract}
 We present highly efficient algorithms for performing forward and backward propagation of Convolutional Neural Network (CNN) for pixelwise classification on  images. For pixelwise classification tasks, such as image segmentation and object detection,
surrounding image patches are fed into CNN for predicting the classes of  centered pixels via forward propagation and for updating  CNN parameters via backward propagation. 
However, forward and backward propagation was originally designed for whole-image classification. Directly applying it to pixelwise classification in a patch-by-patch scanning manner is extremely inefficient, because surrounding patches of pixels have large overlaps, which lead to a lot of redundant computation.

The proposed algorithms \textbf{eliminate all the redundant computation} in  convolution and pooling on images by introducing novel $d$-regularly sparse kernels. \textbf{It generates exactly the same results as those by  patch-by-patch scanning}. Convolution and pooling operations with such kernels are able to continuously access memory and can run efficiently on GPUs. A fraction of patches of interest can be chosen from each training image for backward propagation by applying a mask to the error map at the last CNN layer. Its computation complexity is constant with respect to the number of patches sampled from the image. Experiments have shown that our proposed algorithms speed up commonly used patch-by-patch scanning over \textbf{1500 times} in both forward and backward propagation. The speedup increases with the sizes of images and patches. 
Source code of  GPU implementation is ready to be released to the public. 
\end{abstract}

\section{Introduction}

Convolutional Neural Networks (CNNs) are trainable multistage feed-forward neural networks \cite{lecun_IEEE_98}. They have been extensively investigated to extract good hierarchical feature representations for  image recognition tasks. 
CNN includes three types of layers: convolution layer, pooling layer and non-linearity layer. 
The input and output of each layer are called \emph{feature maps}. 

{\bf The convolution layer} convolves input feature maps with 3D filter banks to generate output feature maps. Each filter extracts the same type of local features at all locations of the input feature map. Conventionally, a convolution layer has a stride of 1\footnote{A stride of $d$ in convolution and pooling layers denotes that  the centers of every two neighboring patches extracted from the input feature map are exactly $d$-pixel away from each other. It down-samples the output feature map such that its height and width are of $1/d$ of the original values.}. But in some recent CNN models, greater-than-1 strides were also used in convolution to down-sample the output feature map.

{\bf The pooling layer} decreases the resolution of the feature maps to make the output feature maps less sensitive to  input shift and distortions. 
Max-pooling and average-pooling are most commonly used. Conventionally, a pooling layers has a stride equalling its kernel size. But in some recent CNN models, strides different than kernel sizes were also used.

{\bf The non-linearity layer} is a point-wise non-linear function applied to each entry of  feature maps.

\begin{figure*}[t]
    \centering
    \begin{tabular}{c@{\hspace{0mm}}c}
        &
        \includegraphics[scale=0.6]{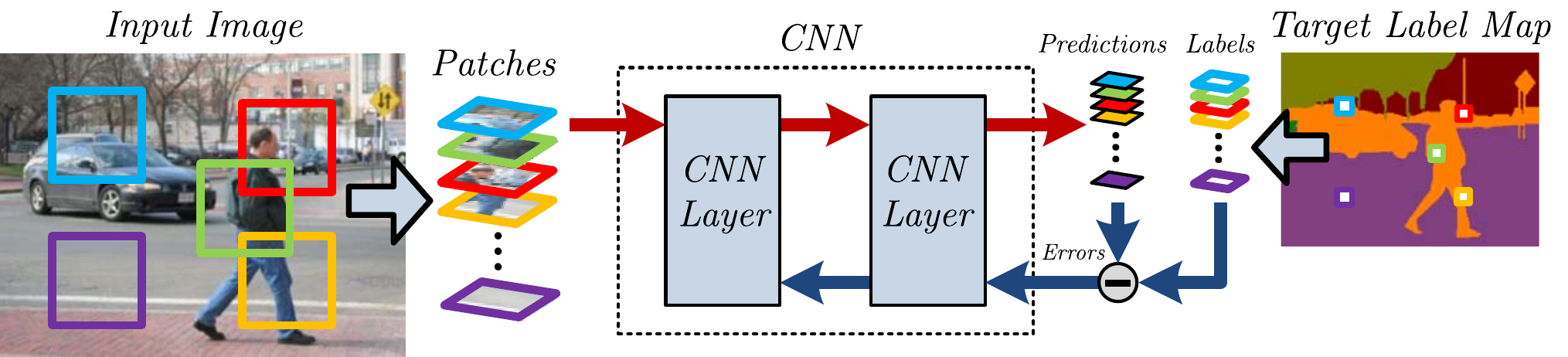}\\
        &(a) Patch-by-patch scanning for CNN based pixelwise classification\\
        &\includegraphics[scale=0.6]{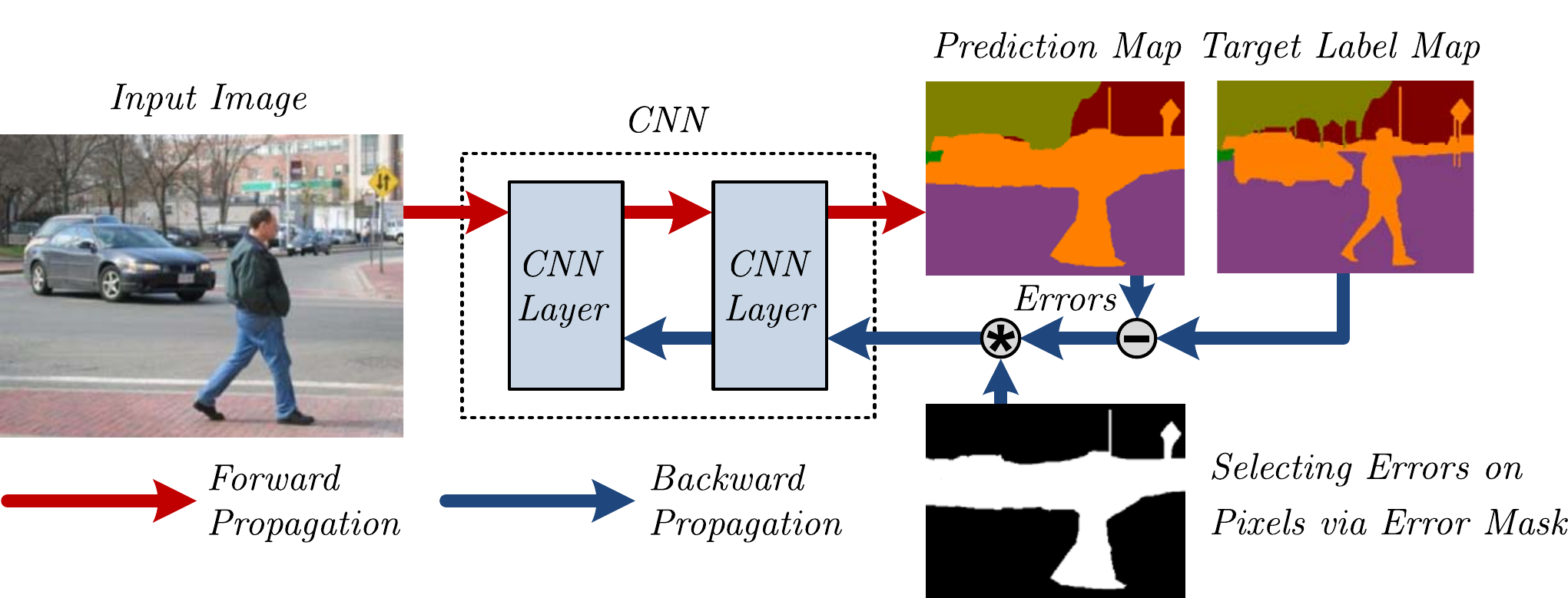}\\
        &(b) Our  approach
    \end{tabular}
    \caption{Comparison of  (a)  patch-by-patch scanning  and (b) the proposed efficient forward and backward propagation for pixelwise classification. The scene labeling task is used for illustration here.}
    \label{fig:intro}
\end{figure*}

After extracting features with a multilayer convolutional network, fully connected layers with a final classifier are added to output class predictions. Given training samples and their labels, the parameters of CNNs are learned in an end-to-end supervised way by minimizing a loss function on  training data.
Forward and backward propagation is used to make class predictions for input samples and to update  CNN parameters based on  prediction errors, respectively.

CNN together with its forward and backward propagation algorithms was originally designed for whole-image classification, i.e., predicting one label for a whole image. 
CNN-based OCR algorithms \cite{lecun_IEEE_98}, \cite{simard_icdar_03}, \cite{chellapilla_IWFHR_06}, \cite{sermanet_icpr_12} drew a lot of attention and were improved over the last decade. 
With deep CNN, Krizhevsky et al. \cite{krizhevsky_nips_06} won the image classification challenge in ImageNet LSVRC 2012 and beat other computer vision algorithms with  large margins. 
\textit{In all the applications above, the input samples of CNNs are whole images without redundant information between them, and therefore they can be processed independently.}

In recent years, CNN has also been applied to object detection \cite{rowley_pami_98}, \cite{frome_iccv_09}, \cite{sermanet_cvpr_13}, \cite{overfeat}, \cite{oquab_cvpr_14}, image segmentation \cite{wu_icpr_14}, scene labeling \cite{farabet_pami_13}, \cite{pinheiro_icml_14}, and tracking \cite{cnn_tracking_10}, and significantly improved the accuracies in these applications. \textit{These tasks are considered as pixelwise classification, i.e., predicting a class label for every pixel, and have fundamental difference with whole-image classification problems}. The input samples of CNNs are image patches surrounding pixels and have large overlaps. Studies \cite{farabet_pami_13}, \cite{pinheiro_icml_14} have shown that inputting larger image patches to CNNs leads to better accuracies, since CNNs can capture more contextual information. In \cite{pinheiro_icml_14}, the chosen patch size covers 1/3 of the whole image. However, this implies larger overlaps between patches. 

Existing approaches ignored such difference and still process image patches independently in a way like treating whole-images and without modifying the forward and backward propagation algorithms. They involved a lot of redundant computation on overlapped patches, and the redundancy increases with both image size and patch size. Figure \ref{fig:intro}.(a) shows straightforward patch-by-patch scanning for both forward and backward propagation. Computation efficiency has become a major bottleneck for these CNN based pixelwise classification tasks. As a compromised solution, one could sacrifice the resolution of the predicted label maps by subsampling, such that overlaps between image patches can be reduced. In object detection, some image patches can be early rejected by fast algorithms before being feed to CNNs, however, sacrificing recalls. Even given that, redundancy still exists and many CNNs based approaches for pixelwise classification were not suitable for realtime applications. 

In pixelwise classification tasks, it is easy to collect thousands of training image patches from a single image, since every pixel has a label. From a large image set, the number of available training samples could reach one billion. Existing approaches treated these training patches independently. Due to the efficiency problem, it is impossible to make use of all the available training samples. Usually only a small subset was randomly sampled for training.

\subsection{Our approach}

In this paper, we propose highly efficient forward and backward propagation algorithms for CNN based pixelwise classification. It generates exactly the same result as  patch-by-patch scanning, without sacrificing the resolutions of predicted label maps or early rejecting any patches. Experimental results show that more than $1,500$ times speedup is achieved on $256  \times 256$ images for both forward and backward propagation. Theoretical analysis shows that compared with patch-by-patch scanning, the complexity of our algorithms is much less sensitive to patch size and the speedup increases with the sizes of images and patches. This is important since image resolutions will become higher and higher in future applications and CNNs prefer large patches which contain more contextual information.

The proposed algorithms also have  potential impact on CNN training. With fast forward propagation, the prediction errors of all the pixels in an image can be estimated quickly at every backward propagation iteration. As shown in Figure \ref{fig:intro}.(b), based on the error map, an arbitrary portion of pixels of interest (even all) on an image can be selected by a mask, and their surrounding patches are used to update CNN with our modified backward propagation. The computation complexity of our backward propagation is constant with respect to the number of image patches selected. 

Figure \ref{fig:intro} compares patch-by-patch scanning and our approach. At the test stage, patch-by-patch scanning sequentially and independently feeds  patches to CNN and the forward propagation is repeated for all the pixels. In our approach, the whole image is taken as the input of CNN which predicts the whole label map with only one pass of the modified forward propagation. At each training iteration, existing approaches predict the error of each sampled patch and use it to calculate gradients with backward propagation. If a mini-batch contains $K$ training patches, both forward propagation and backward propagation are repeated for $K$ times and the gradients estimated from the $K$ patches are averaged to update CNN parameters.
In our approach, a whole-image and its label map are treated as an input-target pair. With the proposed fast forward propagation, class labels at all the pixels can be quickly predicted, and all the prediction errors in the same image can be used to update CNN parameters with only one pass of the modified backward propagation.      


If with 1-stride convolution and without pooling layers, it is not difficult to implement the one-pass forward propagation and one-pass backward propagation described above. Otherwise it is nontrivial, because convolution and pooling operations with greater-than-1 strides have down-sampling effect within each patch. The key of our approach is to modify both the convolution and pooling kernels of the original CNN by inserting a specific number of all-zero columns and rows to compensate for the down-sampling by the convolution and pooling layers. We call such kernels the $d$-regularly sparse kernels. Moreover, based on $d$-regularly sparse kernels, all strides of the convolution and pooling operations become 1, regardless of the strides of convolution and pooling in the original CNN. The 1-strides ensure continuous memory access, which is the key to maximize the computational efficiency on GPUs.


The main contributions of this work can be summarized as three-fold. (1) Our proposed algorithms eliminate all the redundant computation of forward and backward propagation in CNN based pixelwise classification, and achieve significant speedup.  (2) The proposed $d$-regularly sparse kernels not only ensure exactly the same results as patch-by-patch scanning in both forward and backward propagation, but also allow to access memory in a continuous manner, which is the key to fully utilize the computational capability of GPUs, regardless of the strides of convolution and pooling in the original CNN. 3) By applying a mask to the error map at the last layer of a CNN, one can choose an arbitrary subset of patches of interest from a training image to update CNN parameters via backward propagation and with constant computation complexity.



\section{Related Work}

 There have been previous works \cite{krizhevsky_nips_06}, \cite{dean_nips_12}, \cite{mathieu_fft_14} on efficient computation of CNNs. But most methods assumed input samples are independent and did not take the redundant computation between  samples into account. 

Our work is most related to the fast scanning method in \cite{giusti_icip_13}, which was applied in scene labeling \cite{pinheiro_icml_14}. Fast scanning  can be viewed as performing convolution or pooling with different starting offsets to generate multiple feature ``fragments''. The output fragments at the last layer were re-organized to generate the final label maps. Compared with fast scanning \cite{giusti_icip_13}, our key advantage is to ensure 1-strides in all convolution and pooling operations after introducing the $d$-regularly sparse kernels. It allows to access memory addresses continuously, which is the key to fully utilize the computational power of GPUs. In contrast, fast scanning \cite{giusti_icip_13} still keeps large strides in its operations and is not ideal for GPU implementation. It was only implemented on CPUs in \cite{giusti_icip_13}, \cite{pinheiro_icml_14}. Even with GPU implementation, it is multiple times slower than our algorithms.

There are works \cite{luo_cvpr_14} taking whole images as inputs to perform forward and backward propagation with multilayer fully connected networks. However, these works used fully connected layers and are therefore only suitable for images with regular structures such as pedestrians or faces, but not for general objects and images.



\section{Efficient forward propagation}
\subsection{Convolutional neural network}

For CNN consisting of $K$ layers, without loss of generalization, we assume that the input and output of each layer consist of only one 2D feature map throughout the paper. Let $x_k \in {\bf R}^{n_1\times n_2}$ and $y_k \in {\bf R}^{m_1\times m_2}$ denote the input   and  output feature map of the $k$th layer, respectively. $y_k$ is also the input of the next layer, i.e., $y_k = x_{k+1}$. 

If the $k$th layer is a convolution layer, we denote $W_k \in {\bf R}^{l_k\times l_k}$ and $b_k \in {\bf R}$ as the convolution kernel and the bias parameter of this layer. The output of this layer is  
$y_k = W_k *^{d_k} x_k + b_k$,
 where $*^{d_k}$ denotes the convolution operation on $x_k$ with a kernel $W_k$ and a stride of $d_k$.

If the $k$th layer is a pooling layer, it can be viewed as first extracting feature patches from strided locations of the input feature map with a binary mask $P_k \in {\bf R}^{p_k\times p_k}$. The maximal or the mean value of each feature patch is then calculated as the pooling result of that patch. Let $y_k = x_k \circledcirc^{d_k}  P_k$ denote a general pooling operation with a binary kernel $P_k$ and a stride of $d_k$ on the input feature map.

The parameters of a $K$-layer CNN can be optimized by gradient descent. For pixelwise classification tasks, patches centered at pixels of  training images are cropped  as  training samples. For each patch in an input image $I$, the CNN outputs a prediction or a score.
When feeding the image $I$ as the input of the CNN by setting $x_1 = I$, forward propagation cannot generate a prediction at every pixel location due to the greater-than-1 strides in convolution and pooling layers.

Our goal is to make a one-time scan for each layer and generate exactly the same result as the time-consuming patch-by-patch scanning, however, without redundant computation. To achieve this goal, we introduce  $d$-regularly sparse kernels to replace  convolution and pooling kernels. Then it allows to take a whole image as input and directly output a label map for all the pixels without loss of resolution.

 \begin{figure}[t]
    \centering
    \small
    \begin{tabular}{c@{\hspace{0mm}}c@{\hspace{8mm}}c}
        &
        \begin{tabular}{c}
            \includegraphics[scale=0.5]{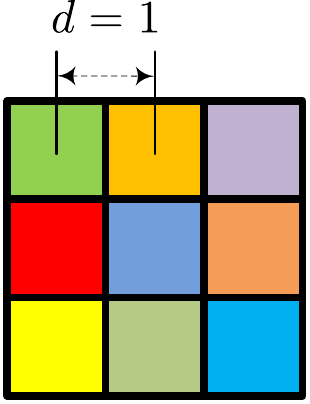}
        \end{tabular}
        &
        \begin{tabular}{c}
            \includegraphics[scale=0.5]{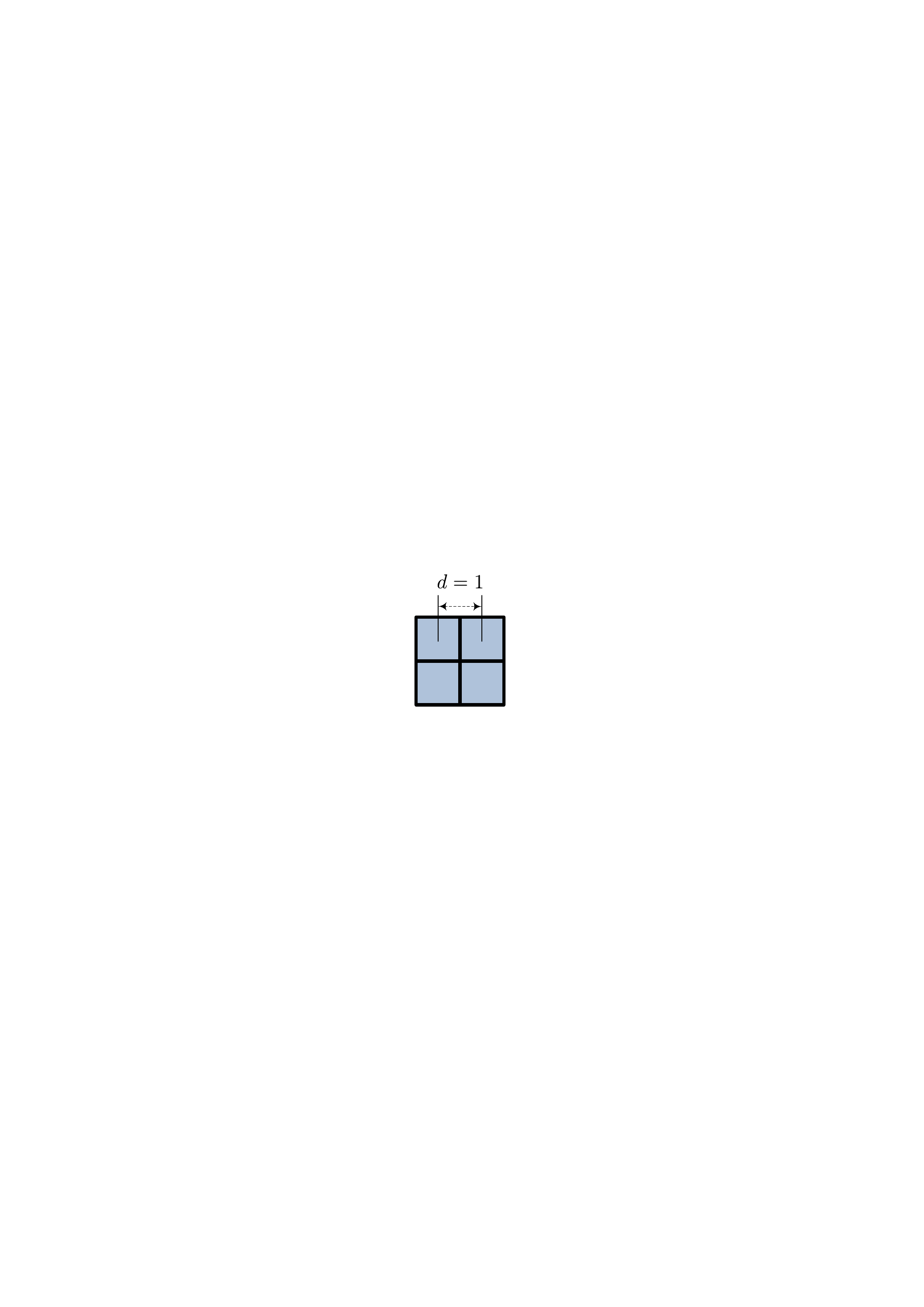}
        \end{tabular}\\
        &(a) $W_k \in {\bf R}^{3\times 3}$ & (b) $P_k \in {\bf R}^{2\times 2}$ \\
        &
        \begin{tabular}{c}
            \includegraphics[scale=0.5]{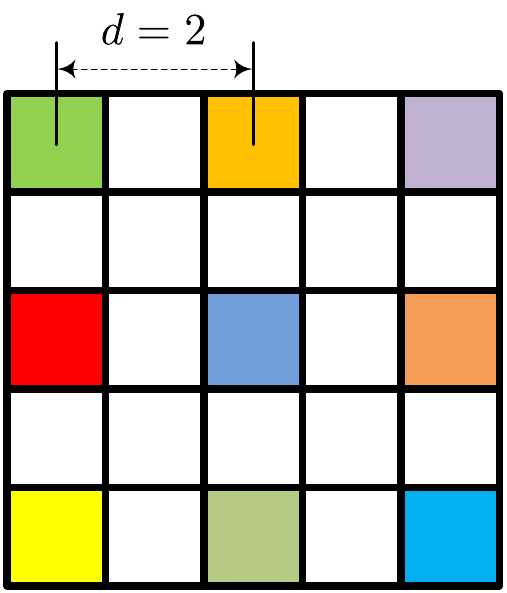}
        \end{tabular}
        &
        \begin{tabular}{c}
            \includegraphics[scale=0.5]{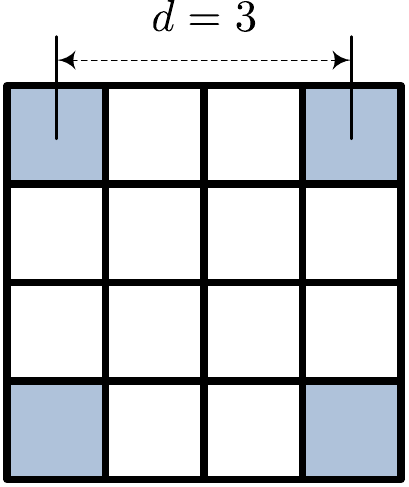}
        \end{tabular}\\
        &(c) $W_{k}$ conversion result & (d) $P_{k}$ conversion result
    \end{tabular}
    \caption{(a) $3\times3$ convolution kernel $W_k$ whose entries are generally non-zeros (represented by colored squares). (b) $2\times2$ pooling kernel $P_k$ whose entries act as binary masks to extract features only at masked locations (represented by shaded squares). (c) Convert the convolution kernel $W_k$ in (a) to a $2$-regularly sparse convolution kernel $W_{k,2}$. Colored squares represent entries from kernel $W_k$, and white squares represent 0. (d) Convert the pooling kernel $P_k$ in (c) to a $3$-regularly sparse pooling kernel $P_{k,3}$. Shaded (white) squares represent masked (unmasked) locations.}
    \label{fig:sparsekernel}
\end{figure}

\subsection{$d$-regularly sparse kernels}
\label{ssec:sparsekernel}

For  convolution kernel $W_k$ and pooling kernel $P_k$, each entry is 1-pixel away from its neighboring entries (see Figures \ref{fig:sparsekernel}.(a) and \ref{fig:sparsekernel}.(b) for examples). We create $d$-regularly sparse kernels for convolution and pooling layers by intersting all-zero rows and columns into the original kernels to make every two original neighboring entries $d$-pixel away. The $d$-regularly sparse convolution and pooling kernels are denoted by $W_{k,d}$ and $P_{k,d}$, respectively. In Figures \ref{fig:sparsekernel}.(c) and \ref{fig:sparsekernel}.(d), we show a 2-regularly sparse kernel $W_{k,2}$, and a 3-regularly sparse kernel $P_{k,3}$. The original kernels can be viewed as 1-regularly sparse kernels where $d=1$, i.e., $W_k = W_{k,1}$ and $P_k = P_{k,1}$. Note that the $d$-regularly sparse kernels are not equivalent to using a stride of $d$ in the conventional convolution and pooling layers. Actually, with our $d$-regularly sparse kernels, all strides in the convolution and pooling layers of the modified CNN are fixed to 1.

\subsection{Image padding}
\label{ssec:imagepadding}

Note that the original CNN is trained with image patches centered at pixel locations of training images. If a CNN takes $n \times n$ image patches as inputs, a test image of size $h \times w$ should be padded to the size of $(h+2\lfloor n/2 \rfloor) \times (w+2\lfloor n/2 \rfloor)$ to ensure patches centered at the border of the original image are also of size $n \times n$. 

\begin{figure*}[t]
    \centering
    \footnotesize
    \begin{tabular}{c}
        \begin{tabular}{c@{\hspace{8mm}}c}
            \includegraphics[scale=0.45]{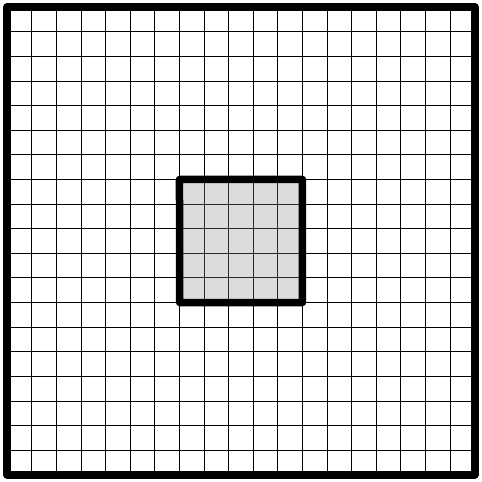}&
            \includegraphics[scale=0.45]{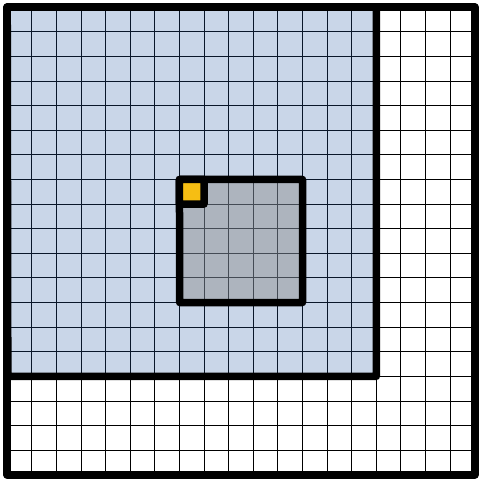}\\
            (a) Padded Input Image $x_1$ of size $19\times 19$ & (b) Top-left Image Patch $\tilde{x}_1$ of size $15\times 15$
        \end{tabular}\\
        \begin{tabular}{c@{\hspace{-7mm}}c@{\hspace{0mm}}c@{\hspace{0mm}}c@{\hspace{0mm}}c}
            &
            \begin{tabular}{c}
                \includegraphics[scale=0.4]{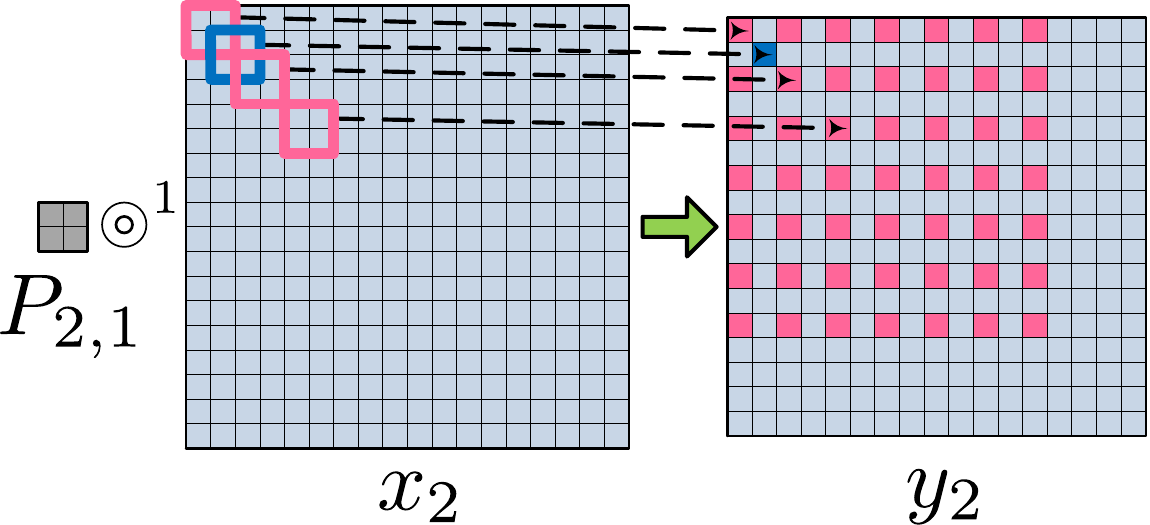}\\
            \end{tabular}
            &
            \begin{tabular}{c}
                \includegraphics[scale=0.4]{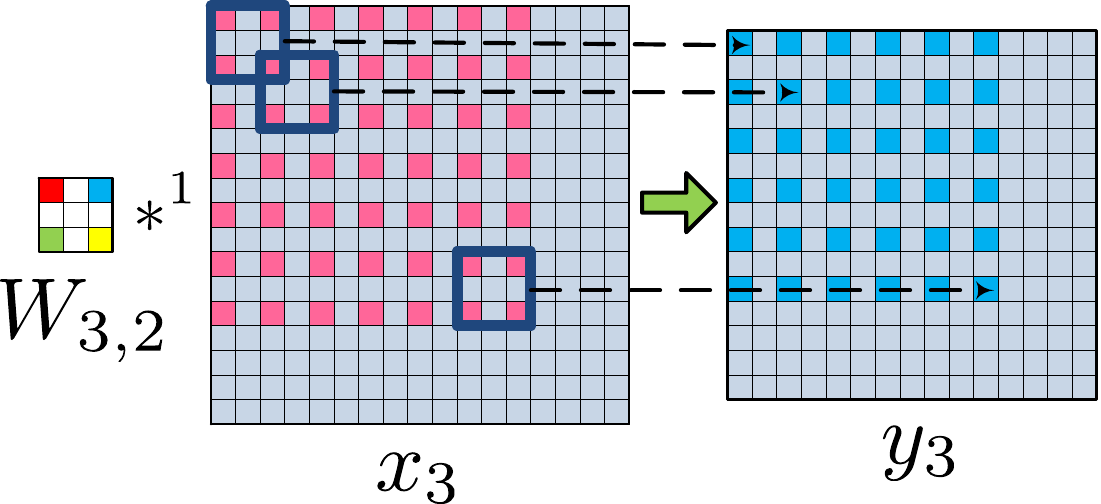}
            \end{tabular}
            &
            \begin{tabular}{c}
                \includegraphics[scale=0.4]{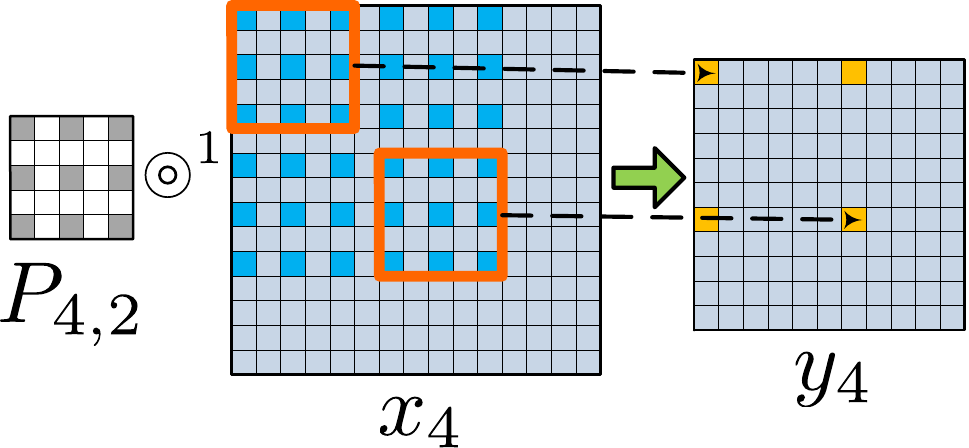}
            \end{tabular}
            &
            \begin{tabular}{c}
                \includegraphics[scale=0.4]{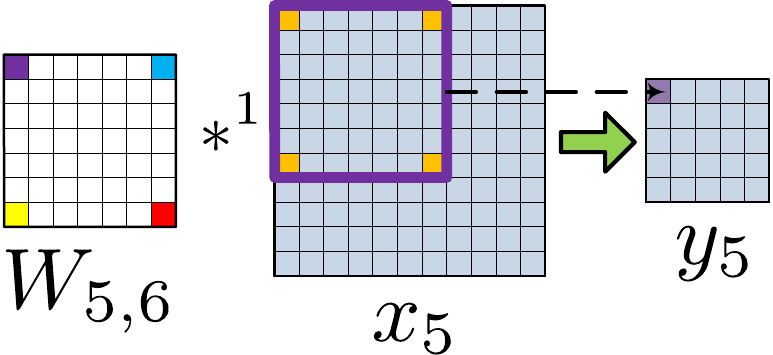}
            \end{tabular}\\
            & (c1) Proposed Pooling & (d1) Proposed Convolution & (e1) Proposed Pooling & (f1) Proposed Convolution\\
            & $P_{2,1} \circledcirc^1 x_2$ & $W_{3,2} *^1 x_3$ & $P_{4,2} \circledcirc^1 x_4$ & $ W_{5,6} *^1 x_5$\\
            &
            \begin{tabular}{c}
                \includegraphics[scale=0.4]{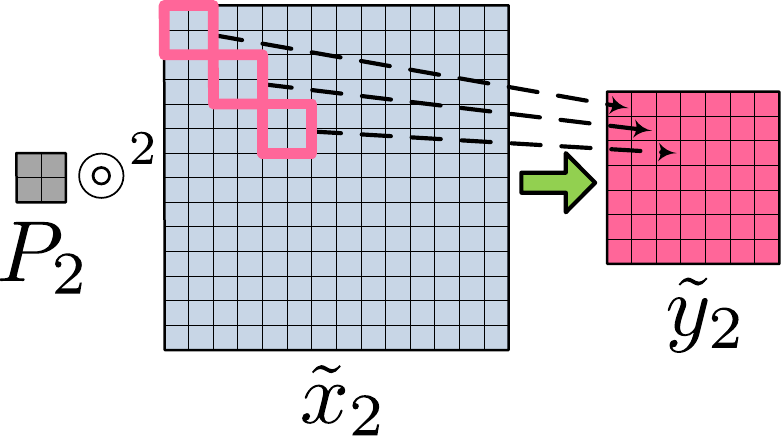}
            \end{tabular}
            &
            \begin{tabular}{c}
                \includegraphics[scale=0.4]{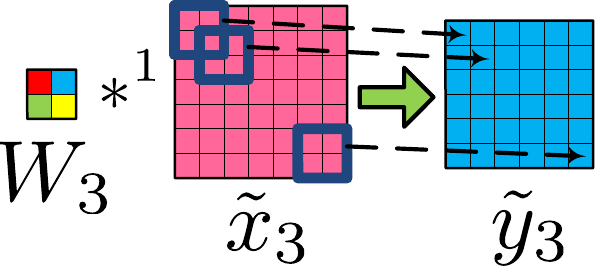}
            \end{tabular}
            &
            \begin{tabular}{c}
                \includegraphics[scale=0.4]{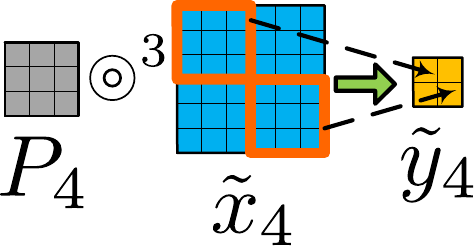}
            \end{tabular}
            &
            \begin{tabular}{c}
                \includegraphics[scale=0.4]{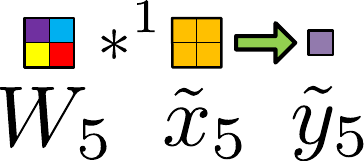}
            \end{tabular}\\
            & (c2) Original Pooling & (d2) Original Convolution & (e2) Original Pooling & (f2) Original Convolution\\
            & $P_{2} \circledcirc^2 \tilde{x}_2$ & $W_{3} *^1 \tilde{x}_3$  & $P_4 \circledcirc^3 \tilde{x}_4$  & $W_5 *^1 \tilde{x}_5$
        \end{tabular}
    \end{tabular}
    \caption{Illustration of our  forward propagation using a CNN with 3 convolution and 2 pooling layers described in Section \ref{ssec:efficientfp}.  (a) A $5\times 5$ input image (shown as shaded squares) is padded to  $19\times19$ to ensure a patch centered on the border of the original image can cover $15\times15$ pixels. In our algorithm, the padded  image is treated as the input of  CNN and  denoted as $x_1$. (b)  $15\times15$  patch (shown as blue squares) centered at the top-left   (shown as the yellow square) of the $5\times5$ image. In the original CNN, the  patch is cropped and fed to the network as input $\tilde{x}_1$. (c1)  Pooling on $x_2 \in {\bf R}^{18\times18}$ with  pooling kernel $P_{2,1} \in {\bf R}^{2\times2}$. $x_2$ is obtained by convolving $x_1$ in (a) with the convolution kernel $W_{1,1}=W_1$. The pooling result is a $17\times17$ feature map $y_2$. Red squares represent feature scores corresponding to those obtained by the original CNN on $\tilde{x}_2$ after pooling (as shown in (c2)). (c2) Pooling in the original CNN on $\tilde{x}_2 \in {\bf R}^{14\times14}$. $\tilde{x}_2$ is obtained by convolving $\tilde{x}_1$ in (b) with the convolution kernel $W_1$. The pooling result is a $7\times7$ feature map $\tilde{y}_2$. (d1) Convolving the feature map $x_3 \in {\bf R}^{17\times17}$ with the regularly sparse convolution kernel $W_{3,2}$. The result is a $15\times15$ feature map. Blue squares represent feature scores corresponding to those obtained by convolving $W_3$ with $\tilde{x}_3$ in the original CNN (as shown in (d2)). (d2) Convolving the feature map $\tilde{x}_3$ with original convolution kernel $W_3$. The result is a $6\times6$ feature map $y_3$. (e1)  Pooling on $x_4 \in {\bf R}^{15\times15}$ with regularly sparse pooling kernel $P_{4,2}\in {\bf R}^{5\times5}$. The result is a $11\times11$ feature map $y_4$. Yellow squares represent feature scores corresponding to those obtained with the original CNN on $\tilde x_4$ after pooling (as shown in (e2)). (e2)  Pooling with the original CNN on $\tilde{x}_4 \in {\bf R}^{6\times6}$. The result is a $4\times4$ feature map $\tilde{y}_4$. (f1) Convolving the feature map $x_5 \in {\bf R}^{11 \times 11}$ with the regularly sparse convolution kernel $W_{5,6}$. The result is a $5\times5$ feature map $y_5$. Purple squares represent feature scores corresponding to those obtained by convolving $W_5$ with $\tilde{x}_5$ in the original CNN. (f2) Convolving the feature map $\tilde{x}_5$ with original convolution kernel $W_5$, which results in a singe feature score $\tilde{y}_5$.}
    \label{fig:overall}
\end{figure*}

 \begin{algorithm}[tb]
    \small
    \caption{Efficient Forward Propagation of CNN}
    \label{alg:overall}
    \KwIn{Input image $I$, convolution parameters $W_k$, $b_k$, pooling kernels $P_k$, strides of each layer $d_k$}
    \Begin{
        $d \leftarrow 1$\\
        $x_1 \leftarrow I$\\
        \For{$k = \{1,2,\cdots,K\}$}
        {
            \uIf{Layer $k$ == convolution layer}
            {
                 Convert $W_k$ to $W_{k,d}$ (see Section \ref{ssec:sparsekernel})\\
                 $y_k \leftarrow W_{k,d} *^{1} x_k + b_k$,
            }
            \ElseIf{Layer $k$ == pooling layer}
            {
                 Convert $P_k$ to $P_{k,d}$ (see Section \ref{ssec:sparsekernel})\\
                 $y_k \leftarrow P_{k,d} \circledcirc^{1} x_k$
            }
            $d \leftarrow d \times d_k$\\
            $x_{k+1} \leftarrow y_k$\\
        }
        \Return{output feature map $y_{k}$}\\
        
    }
 \end{algorithm}

\subsection{Efficient forward propagation}
\label{ssec:efficientfp}

Our  algorithm for efficient forward propagation is summarized in Algorithm \ref{alg:overall}. Note that strides in all layers are fixed to 1. We explain the algorithm step-by-step by converting a conventional CNN which includes 3 convolution layers and 2 max-pooling layers to a new one with regularly sparse kernels. For simplicity, non-linearity layers are not included and only strides of  pooling layers are greater than 1. The original CNN is composed of a $2\times2$ convolution layer followed by a $2\times2$ pooling layer with a stride of $2$, another $2\times2$ convolution layer and a $3\times3$ pooling layer with a stride of $3$, and a final $2\times2$ convolution layer to generate a feature score. The original network takes $15\times 15$ image patches as inputs and outputs a single feature score after the forward propagation. The output feature map of the CNN is of size $1\times1$.
Given a $5\times 5$ input image\footnote{We choose an image size smaller than the patch size for the convenience of illustration; otherwise, the figure will be too big.}
 (Figure \ref{fig:overall}.(a)), where each pixel needs a prediction score, it is first padded to $19\times19$. The image patch centered at the top-left corner of the original image is shown in Figure \ref{fig:overall}.(b).

We illustrate how our algorithm computes at each layer of the modified CNN in Figure \ref{fig:overall} and compare its computation with that of the original CNN by using the full input image in Figure \ref{fig:overall}.(a) and the top-left image patch in Figure \ref{fig:overall}.(b). For the first convolution and pooling layers, the proposed algorithm performs  convolution in the same way as the original CNN, but pools $2\times2$ patches with stride $d_2=1$. The difference between our  algorithm and the original CNN for the first pooling layer is illustrated in Figure \ref{fig:overall}.(c). 
Our  algorithm performs convolution on the whole padded input image and does not reduce its resolution after pooling with stride $d_2=1$. In contrast, the original CNN performs convolution only on one patch and the resolution of the feature map is reduced after pooling.
For our modified CNN, because the stride in the previous pooling layer equals 1, the input feature maps for the second convolution and pooling layers are not equivalent to the ones obtained by the original CNN. As shown in Figures \ref{fig:overall}.(d)-\ref{fig:overall}.(e), each pair of neighboring entries obtained with the original CNN is now $2$-pixel away from each other in the feature map obtained with our algorithm.
To generate same feature scores as the original CNN after another round of convolution and pooling operations, the convolution kernel $W_3$ and pooling kernel $P_4$ should be converted to 2-regularly sparse kernels $W_{3,2}$ and $P_{4,2}$ to let neighboring entries in these kernels be $2$ pixels away. After the 2nd pooling layer in our modified CNN, each pair of neighboring entries in the output feature map obtained with the original CNN is now $d_2\times d_4=6$ pixels away from each other in the output feature map. Therefore, the convolution kernels $W_5$ should be converted to $6$-regularly sparse kernels $W_{5,6}$ to generate the final feature map (see Figure \ref{fig:overall}.(f)).

Since fully connected layers can be viewed as convolution operations, such layers following the convolution or poolying layers in the original CNN can also be converted to convolution layers with regularly sparse kernels.

\subsection{Theoretical speedup of forward propagation}
\label{ssec:speedup}

We assume that each convolution layer is followed by a pooling layer and a non-linearity layer. Let $s^2$ be the number of pixels in the input feature map of the 1st layer (usually a pre-processed image), $m_k^2$ be the number of pixels in each image patch at layer $k$, $l_k^2$ be the number of pixels of the convolution kernel at layer $k$, and $p_k^2$ be the number of foreground masks in the pooling kernel following layer $k$. The computation complexity of patch-by-patch scanning at layer $k$ with a stride of 1 can be calculated as,
  \begin{align}
        O\big( s^2 \cdot(m_k^2 l_k^2 + 2m_k^2) \big) \approx  O\big( s^2 m_k^2 l_k^2 \big).
  \end{align}
On the left-hand side, the $s^2$ term denotes that a total of $s^2$ image patches are evaluated, the $m_k^2 l_k^2$ term denotes the complexity of convolving each image patch at the $k$th convolution layer, and the $2m_k^2$ term denotes that each pixel is compared or added once at the pooling layer followed by a non-linear transformation at the non-linearity layer.

 For our algorithm, the time complexity is calculated as
 \begin{align}
    O\big( (s+m_k)^2 (l_k^2 + p_k^2 + 1) \big) \approx O\big( (s + m_k)^2 l_k^2 \big).
 \end{align}
On the left-hand side, the $(s+m_k)^2$ term denotes that the input image needs to be padded before being processing as described in Section \ref{ssec:imagepadding}, $l_k^2$ denotes the complexity of convolving the input feature map at layer $k$, $p_k^2$ denotes the complexity of pooling following layer $k$, and $1$ denotes the complexity of applying point-wise non-linear transformation to output feature maps. On the right hand side, $p_k^2$ is omitted because it is usually smaller than $l_k^2$, and pooling operations are generally much faster than convolutions.

 Our algorithm has a speedup of $O( s^2 m_k^2 / (s + m_k)^2 )$ compared with the patch-by-patch scanning. 
 The speedup increases with image size and image patch size. 
 Since the size of intermediate feature map $m_k^2$ gradually decreases due to greater-than-1 strides, the speedup is the largest for the $1$st layer and gradually decreases. 

\begin{figure}[t]
    \centering
    \begin{tabular}{c@{\hspace{-1mm}}c}
        &
        \begin{tabular}{c}
        	\includegraphics[scale=1]{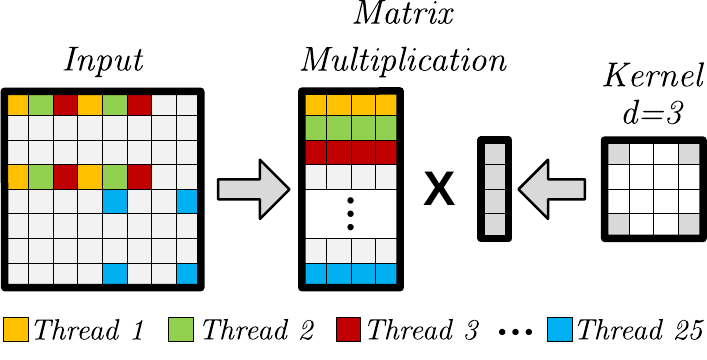}\\
        	(a) Memory accessed by 25 threads
        \end{tabular}\\
        &
        \begin{tabular}{c@{\hspace{2mm}}c@{\hspace{6mm}}c@{\hspace{2mm}}c}
        	\includegraphics[scale=0.8]{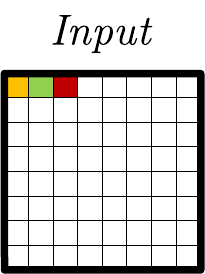}&
           \includegraphics[scale=0.8]{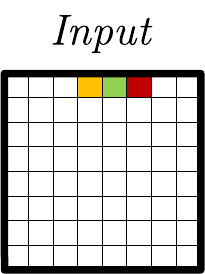}&
           \includegraphics[scale=0.8]{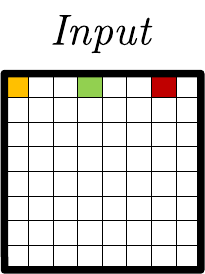}&
           \includegraphics[scale=0.8]{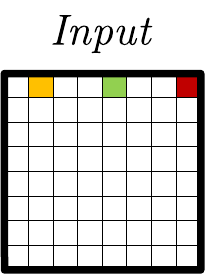}\\
           (b) Iter. 1 & (c) Iter. 2 & (d) Iter. 1 & (e) Iter. 2 \\
           by ours & by ours & by \cite{giusti_icip_13} & by \cite{giusti_icip_13}
        \end{tabular}
    \end{tabular}
    \caption{(a) Input feature map is accessed iteratively by our proposed convolution operation with 25 GPU threads. Each thread extracts 4 values iteratively from the input feature map to form a matrix. Convolution is performed by matrix multiplication with the original kernel. (b)-(c) GPU threads 1-3 in (a) access the input feature map at iterations 1 and 2. Note that the memory addresses accessed by the threads are consecutive. (d)-(e) Illustration of how GPU threads 1-3 access strided locations of the GPU memory at iterations 1 and 2 by fast scanning in \cite{giusti_icip_13}. }
    \label{fig:implementation}
\end{figure}

\subsection{GPU implementation}
\label{ssec:implementation}

Our algorithm can run very efficiently on GPU. The efficiency on GPU is  limited by the way it accesses  GPU memory. Our  forward propagation algorithm  has the advantage of continuously accessing  GPU memory by threads in the same blocks, which is the key to fully utilize the computational capacity of  GPU.


The Caffe library \cite{jia_caffe_14} provides one of the fastest implementations of  convolution on GPU \cite{convnets_benchmarks}. 
It extracts feature patches from the input feature map and convert them into a large matrix.  Convolution is  calculated as a matrix multiplication between this large matrix and the convolution kernel.
Our algorithm is implemented based on the \texttt{conv\_layer} of the Caffe library. Every thread is in charge of extracting values from the input feature map for calculating one entry on the output feature map. At every location of the input feature map, all values specified by the non-zero entries of the convolution kernel are iteratively extracted by a single thread. The extracted values are then organized into a large matrix and multiplied by the convolution kernel to generate the final result (Figure \ref{fig:implementation}.(a)). In this way, consecutive threads in a same block can access consecutive addresses in the GPU memory and take full use of the GPU memory bandwidth (Figures \ref{fig:implementation}.(b) and \ref{fig:implementation}.(c)).

The max and average pooling can be implemented in a similar way, i.e., 
each GPU thread performing max or average calculation on the extracted feature patches for one output entry. 
Thus the continuous access to the GPU memory is achieved in both convolution and pooling layers.


Fast scanning \cite{giusti_icip_13} performs convolution or pooling operations with greater-than-1 strides in the original manner but with different starting offsets. 
Therefore, it has to access strided addresses of memory (Figures \ref{fig:implementation}.(d) and \ref{fig:implementation}.(e)), which is unable to fully utilize the bandwidth of GPU memory and significantly hinders its efficiency.
Moreover, each operation with different offsets leads to multiple output sub-feature maps of different sizes, and the number of such sub-feature maps increases exponentially as the number of strided layers increases, further hindering its efficiency.

\section{Efficient backward propagation}


Backward propagation on the modified CNN with regularly sparse kernels can be  performed by directly feeding whole images and their pixelwise label maps as the inputs. Compared with a conventional CNN, there are two differences for performing backward propagation on the modified CNN: 1) the errors at the last layer are no longer single values but the errors of all pixels (or a fraction of chosen pixels) in a training image; and 2) only gradients of the non-zeros entries in the regularly sparse convolution kernels are calculated and only those entries are updated during training.


\subsection{Backward propagation of convolution layers}

Let $\boldsymbol\delta_k$ denote the error map corresponding to the input feature map $x_k$ at layer $k$. 
To compute one entry in the error map $\boldsymbol\delta^k$, one should extract the next layer's errors of units that are connected to the entry of interest in $\boldsymbol\delta_k$, then multiply each of them by the associated weights in the convolution kernel, and sum the weighted errors. The calculation of all entries of the error map $\boldsymbol\delta_k$ can be converted to a convolution operation: $\boldsymbol\delta_k = \textnormal{pad}(\boldsymbol\delta_{k+1}) \ast^1 \textnormal{rot}(W_{k,d})$, where $\textnormal{pad()}$ denotes zero padding the error map $\boldsymbol\delta_{k+1}$, and $\textnormal{rot}()$ denotes rotating  convolution kernel $W_{k,d}$ for 180$^{\circ}$. 

For each non-zero entry $W_{k,d}(i)$ in the convolution kernel $W_{k,d}$, its gradient is calculated as the sum of all the weighted errors that are calculated with the entry. The weights for  errors are determined by the input values from the input feature map $x_k$ for convolution: $\nabla W_{k,d}(i) = \sum_{u,v} (\boldsymbol\delta_{k+1})_{u,v} (x_{k})^i_{u,v}$, where $(x_k)^i_{u,v}$ are the input values in $x_k$ and are multiplied elementwise by $W_{k,d}(i)$ during  convolution to compute the entry at $(u,v)$ of the output feature map $x_{k+1}$. Calculating the gradients of  $W_k$ can be converted into a convolution operation: $\nabla W_{k,d} = x_k \ast^1 \textnormal{rot}(\boldsymbol\delta_{k+1,d})$, where $\boldsymbol\delta_{k+1,d}$ denotes that the error map $\boldsymbol\delta_{k+1}$ is inserted with all-zero rows and columns as the kernel $W_{k,d}$ does at layer $k$. Similarly, the gradient for the bias $b_k$ is calculated as the sum of all the entries in $\boldsymbol\delta_{k+1}$. The speedup of backward propagation can be derived similarly to that of forward propagation as $O(s^2 m_{k+1}^2/(s+m_{k+1})^2)$.

\subsection{Backward propagation of pooling layers}

For max pooling with regularly sparse kernels, the index within every patch where the maximal value is chosen is recorded during forward propagation. During backward propagation, the errors at $\boldsymbol\delta_{k+1}$ transfer back to the errors at $\boldsymbol\delta_{k}$, and accumulate at the locations specified by those indices. For average pooling, it can be viewed as a mean filtering and calculated similarly as the convolution layer.

\subsection{Selecting pixels of interest}
\label{ssec:choosingpixels}

We can select prediction errors of only a fraction of pixels in a training image for backward propagation. This is achieved by applying a mask on the error map of the last layer, where the prediction errors of pixels of interest are kept and the errors at other entries are set to zero (see ``Error Mask'' in Figure \ref{fig:intro}.(b)). The gradients calculated in this way are exactly the same as those calculated by extracting the image patches centered at the pixels of interest in the training image and feeding them as a mini-batch into the original CNN. 
The computation complexity does not change when different subsets of pixels are chosen. 

This is an important property for tasks such as object detection, where only a small number of positive samples exist in each training image. If image patches at all pixel locations are used for training, the gradients by the positive samples would be overwhelmed by those calculated with the significantly larger number of negative samples. For scene labeling tasks, other strategies of choosing pixels during training might be beneficial for impoving the accuracy.

\begin{table*}[tb]
  \centering
  \scriptsize
  \begin{tabular}{|c|c|c|c|c|c|c|c|c|c|c|c|c|c|c|c|c|c|c|}
  \hline
  Layer Type & conv1 & pool1 & tanh1 & conv2 & pool2 & tanh2 & conv3 & \multirow{3{•}}{*}{Overall} \\ \cline{1-8}
  Kernel Size / Stride & \tiny{$50\times6\times6$} / $1$ & \tiny{$8\times8$} / $8$ & - & \tiny{$50\times3\times3$} / $1$ & \tiny{$2\times2$} & - & \tiny{$32\times7\times7$} / $1$ & \\ \hline
  Patch-by-Patch & \multirow{2}{*}{22983.8} & \multirow{2}{*}{4916.4} & \multirow{2}{*}{73.71} & \multirow{2}{*}{5066.2} & \multirow{2}{*}{46.68} & \multirow{2}{*}{16.76} & \multirow{2}{*}{22134.8} & \multirow{2}{*}{55238.4} \\
  Fwd. Prop. (ms) &  &  &  &  &  &  & & \\ \hline
  Fast Scanning \cite{giusti_icip_13} & \multirow{2}{*}{3.103} & \multirow{2}{*}{68.04} & \multirow{2}{*}{0.518} & \multirow{2}{*}{10.63} & \multirow{2}{*}{2.464} & \multirow{2}{*}{0.386} & \multirow{2}{*}{72.95} & \multirow{2}{*}{158.09} \\
  Fwd. Prop. (ms) &  &  &  &  &  &  & & \\ \hline
  Our Method & \multirow{2}{*}{3.074} & \multirow{2}{*}{6.688} & \multirow{2}{*}{0.526} & \multirow{2}{*}{7.088} & \multirow{2}{*}{1.211} & \multirow{2}{*}{0.395} & \multirow{2}{*}{16.41} & \multirow{2}{*}{35.39} \\
  Fwd. Prop. (ms) &  &  &  &  &  &  &  & \\ \hline
  Speedup by Ours & \multirow{2}{*}{7476.8} & \multirow{2}{*}{735.1} & \multirow{2}{*}{140.1} & \multirow{2}{*}{714.8} & \multirow{2}{*}{38.5} & \multirow{2}{*}{42.4} & \multirow{2}{*}{1348.86} & \multirow{2}{*}{1560.8} \\
  Fwd. Prop. &  &  &  &  &  &  & & \\ \hline
  Patch-by-Patch & \multirow{2}{*}{56992.3} & \multirow{2}{*}{14765.7} & \multirow{2}{*}{64.53} & \multirow{2}{*}{6886.0} & \multirow{2}{*}{186.3} & \multirow{2}{*}{19.8} & \multirow{2}{*}{8285.2} & \multirow{2}{*}{87199.8} \\
  Bwd. Prop. (ms) &  &  &  &  &  &  & & \\ \hline
  Our Method & \multirow{2}{*}{7.42} & \multirow{2}{*}{14.52} & \multirow{2}{*}{0.481} & \multirow{2}{*}{27.11} & \multirow{2}{*}{1.538} & \multirow{2}{*}{0.424} & \multirow{2}{*}{39.78} & \multirow{2}{*}{91.26} \\
  Bwd. Prop. (ms) &  &  &  &  &  &  & & \\ \hline
  Speedup by Ours & \multirow{2}{*}{7680.9} & \multirow{2}{*}{1016.9} & \multirow{2}{*}{134.2} & \multirow{2}{*}{254.0} & \multirow{2}{*}{121.1} & \multirow{2}{*}{46.7} & \multirow{2}{*}{208.3} & \multirow{2}{*}{955.5} \\
  Bwd. Prop. &  &  &  &  &  &  &  & \\
  \hline
  \end{tabular}
  \caption{The layewise timing and speedup results of forward and backward propagation by our proposed algorithm, and the layerwise timing results of forward propagation by the fast scanning method \cite{giusti_icip_13} on the Plain CNN$_1$ model with $3\times388\times388$ images as inputs.}
  \label{tab:timing1}
\end{table*}

\begin{table*}[tb]
  \centering
  \scriptsize
  \begin{tabular}{|c|c|c|c|c|c|c|c|c|c|c|c|c|c|c|c|c|c|c|}
  \hline
   Layer Type & conv11 & pool11 & tanh11 & conv12 & conv13 & conv21 & pool21 & tanh21 \\ \hline
  Kernel Size / Stride & \tiny{$25\times8\times8$} / $1$ & \tiny{$2\times2$} / $2$ & - & \tiny{$50\times8\times8$} / $1$ & \tiny{$32\times1\times1$} / $1$ & \tiny{$25\times8\times8$} / $1$ & \tiny{$2\times2$} / $2$ & - \\ \hline
  Sliding Window & \multirow{2}{*}{39485.6} & \multirow{2}{*}{1960.2} & \multirow{2}{*}{693.0} & \multirow{2}{*}{59017.2} & \multirow{2}{*}{6473.1} & \multirow{2}{*}{63548.4} & \multirow{2}{*}{332.2} & \multirow{2}{*}{98.14} \\
  Fwd. Prop. (ms) &  &  &  &  &  &  & & \\ \hline
  Our Method & \multirow{2}{*}{4.398} & \multirow{2}{*}{0.854} & \multirow{2}{*}{0.337} & \multirow{2}{*}{24.42} & \multirow{2}{*}{2.466} & \multirow{2}{*}{28.90} & \multirow{2}{*}{0.70} & \multirow{2}{*}{0.227} \\
  Fwd. Prop. (ms) &  &  &  &  &  &  & & \\ \hline
  Speedup by Ours & \multirow{2}{*}{8978.1} & \multirow{2}{*}{2295.3} & \multirow{2}{*}{2056.4} & \multirow{2}{*}{2416.8} & \multirow{2}{*}{2631.3} & \multirow{2}{*}{2198.9} & \multirow{2}{*}{474.6} & \multirow{2}{*}{426.7} \\
   Fwd. Prop. &  &  &  &  &  &  & & \\ \hline
   Sliding Window & \multirow{2}{*}{73961.5} & \multirow{2}{*}{10054.8} & \multirow{2}{*}{602.6} & \multirow{2}{*}{146019.3} & \multirow{2}{*}{25206.7} & \multirow{2}{*}{133706.2} & \multirow{2}{*}{1623.8} & \multirow{2}{*}{106.7} \\
  Bwd. Prop. (ms) &  &  &  &  &  &  &  & \\ \hline
  Our Method & \multirow{2}{*}{8.193} & \multirow{2}{*}{1.428} & \multirow{2}{*}{0.282} & \multirow{2}{*}{66.55} & \multirow{2}{*}{6.778} & \multirow{2}{*}{71.69} & \multirow{2}{*}{0.844} & \multirow{2}{*}{0.245} \\
  Bwd. Prop. (ms) &  &  &  &  &  &  & & \\ \hline
  Speedup by Ours & \multirow{2}{*}{9027.4} & \multirow{2}{*}{7041.2} & \multirow{2}{*}{2136.9} & \multirow{2}{*}{2194.1} & \multirow{2}{*}{3718.9} & \multirow{2}{*}{1865.1} & \multirow{2}{*}{1923.9} & \multirow{2}{*}{6627.8} \\
   Bwd. Prop. &  &  &  &  &  &  & & \\ \hline \hline
   Layer Type & conv22 & conv23 & conv31 & pool31 & tanh31 & conv32 & conv33 & \multirow{2}{*}{Overall} \\ \cline{1-8}
  Kernel Size / Stride & \tiny{$50\times8\times8$} / $1$ & \tiny{$32\times1\times1$} / $1$ & \tiny{$25\times8\times8$} / $1$ & \tiny{$2\times2$} / $2$ & - & \tiny{$50\times8\times8$} / $1$ & \tiny{$32\times1\times1$} / $1$  & \\ \hline
  Sliding Window & \multirow{2}{*}{14765.3} & \multirow{2}{*}{2433.4} & \multirow{2}{*}{17059.8} & \multirow{2}{*}{32.15} & \multirow{2}{*}{13.81} & \multirow{2}{*}{17015.4} & \multirow{2}{*}{2069.7} & \multirow{2}{*}{224997.4} \\
  Fwd. Prop. (ms) &  &  &  &  &  &  & & \\ \hline
  Our Method & \multirow{2}{*}{18.98} & \multirow{2}{*}{1.920} & \multirow{2}{*}{20.55} & \multirow{2}{*}{0.488} & \multirow{2}{*}{0.164} & \multirow{2}{*}{10.76} & \multirow{2}{*}{1.080} & \multirow{2}{*}{116.2} \\
  Fwd. Prop. (ms) &  &  &  &  &  &  & & \\ \hline
  Speedup by Ours & \multirow{2}{*}{777.9} & \multirow{2}{*}{1267.4} & \multirow{2}{*}{830.2} & \multirow{2}{*}{65.9} & \multirow{2}{*}{84.2} & \multirow{2}{*}{1581.4} & \multirow{2}{*}{1916.4} & \multirow{2}{*}{1935.6} \\
  Bwd. Prop. &  &  &  &  &  &  & & \\ \hline
  Sliding Window & \multirow{2}{*}{28744.1} & \multirow{2}{*}{8522.3} & \multirow{2}{*}{16727.5} & \multirow{2}{*}{128.358} & \multirow{2}{*}{15.91} & \multirow{2}{*}{8657.7} & \multirow{2}{*}{2793.6} & \multirow{2}{*}{456871.1} \\
  Bwd. Prop. (ms) &  &  &  &  &  &  & & \\ \hline
  Our Method & \multirow{2}{*}{52.35} & \multirow{2}{*}{5.368} & \multirow{2}{*}{50.89} & \multirow{2}{*}{0.630} & \multirow{2}{*}{0.180} & \multirow{2}{*}{29.47} & \multirow{2}{*}{3.117} & \multirow{2}{*}{298.0} \\
  Fwd. Prop. (ms) &  &  &  &  &  &  & & \\ \hline
  Speedup by Ours & \multirow{2}{*}{549.1} & \multirow{2}{*}{1587.6} & \multirow{2}{*}{328.7} & \multirow{2}{*}{203.7} & \multirow{2}{*}{88.4} & \multirow{2}{*}{293.8} & \multirow{2}{*}{896.2} & \multirow{2}{*}{1533.1} \\
  Bwd. Prop. &  &  &  &  &  &  & & \\ \hline
  \end{tabular}
  \caption{The layewise timing and speedup results of the forward and backward propagation by our proposed algorithm on the RCNN$_3$ model with $3\times410\times410$ images as inputs.}
  \label{tab:timing2}
\end{table*}

\section{Experiments}
\label{sec:experiments}

All the experiments are conducted on an NVIDIA K40 GPU. Fast scanning in \cite{giusti_icip_13} and patch-by-patch scanning are used for comparison. Fast scanning only supports forward propagation. All the methods were implemented based on the Caffe library. The actual running time is used to evaluate the efficiency of the compared methods.


\subsection{Running times of practical CNN models}
\label{ssec:practicalmodels}

We tested the running times of forward and backward propagation of two practical CNN models, the Plain CNN$_1$ and the RCNN$_3$ models, for scene labeling from \cite{pinheiro_icml_14}. 
Detailed network structures are recorded in Tables \ref{tab:timing1} and \ref{tab:timing2}. The output feature map is of size $256\times256$ with 32 channels. The input feature map is padded accordingly to $388\times388$ and $410\times410$ for the two models respectively. Note that the running times of the CNN models depend only on the image size and network structures but not on specific feature and kernel values. Random numbers were filled in the input images and the convolution kernels.

As shown by the layerwise and overall timing results, our proposed method achieves a speedup of over 1500 times compared with the traditional patch-by-patch approach. Compared with the fast scanning method \cite{giusti_icip_13}, our proposed algorithm has a speedup over 10 times at the pool1 layer and a speedup over 2 times at the pool2 layer. Since the fast scanning method outputs multiple sub-feature maps at the pool1 and pool2 layers, the large number of sub-feature maps also hinders the efficiency of the conv2 and conv3 layers. Those timing results show that the performance of the fast scanning method decreases significantly as the stride increases and it is therefore not suitable for GPU implementation.
We also observed that some pooling layers might take even longer time to calculate than the convolution layers. Since the pooling operations are mostly limited by the bandwidth of GPU memory, this again shows the importance of continuous memory access by our proposed algorithm.

We also tested the running time of backward propagation by randomly choosing 128, 512, or 1024 pixels from the error map at the last CNN layer as described in Section \ref{ssec:choosingpixels}. The running times of backward propagation with error masks have no difference with those without masks. 

The maximal numerical differences between the gradients calculated by our proposed algorithm and by the patch-by-patch method are smaller than $10^{-6}$ for all above testing cases. The numerical results validate the correctness of our algorithm and our implementation.

\begin{table}
 \centering
 \scriptsize
 \begin{tabular}{|c|c|c|}
   \hline
   Padded image size & Time (ms) & Overall speedup \\
   \hline
   $3\times260\times260$ & 12.57 & 1098.0\\
   \hline
   $3\times388\times 388$ & 35.39 & 1560.8 \\
   \hline
   $3\times644\times644$ & 121.68 & 1815.9\\
   \hline
 \end{tabular}
 \caption{The timing and speedup results by our proposed algorithm on the Plain CNN$_1$ model with different input image sizes.}
 \label{tab:imagesizes}
\end{table}

\begin{table}
 \centering
 \scriptsize
 \begin{tabular}{|c|c|c|c|c|c|}
   \hline
   Input patch & pool1 kernel & Overall & conv1 & pool1 & Overall \\
   size & size / stride & time (ms) & speedup & speedup & speedup \\
   \hline
   $37\times37$ & $2\times2$ / 2 & 24.53 & 2264.0 & 218.0 & 1293.8 \\
   \hline
   $69\times69$ & $4\times4$ / 4 & 26.84 & 3406.9 & 390.7 & 1320.8 \\
   \hline
   $133\times133$ & $8\times8$ / 8 & 35.39 & 7476.8 & 735.1 & 1560.8 \\
   \hline
 \end{tabular}
 \caption{The timing and speedup results by our proposed algorithm on the Plain CNN$_1$ model with modifications to the pool1 layer to take $37\times37$ and $69\times69$ image patches as inputs.}
 \label{tab:imagepatchsizes}
\end{table}

\subsection{Effects of different image and image patch sizes}
\label{ssec:differentsizes}

We also tested the running times of forward propagation of the above mentioned Plain CNN$_1$ model with different image sizes and image patch sizes.

Images of two additional image sizes, $128 \times 128$ and $512 \times 512$, which are padded to $260\times 260$ and $644\times644$ respectively, are fed into the CNN model as the inputs. The timing and speedup results are reported in Table \ref{tab:imagesizes}.

To make the Plain CNN$_1$ model take image patches of sizes different than $133\times133$ as input and still output single values, we adjusted the pooling kernel size of the pool1 layer to $4\times4$ with a stride of $4$ to take $69\times69$ image patches as inputs, and to $2\times2$ with a stride of $2$ to take $37\times37$ image patches as inputs. The timing and speedup results are recorded in Table \ref{tab:imagepatchsizes}. It shows that the speedup of the conv1 and pool1 layers decrease significantly as the image patch size decreases, but the overall speedup only decreases slightly. This is because image patch sizes after the pool1 layer are not significantly changed.

The speedup results of the above two experiments show that the speedup increases as the image size and image patch size increase, which validates our theoretical analysis. 

\section{Conclusions and future works}
\label{sec:conclusions}
This work has fundamental contributions to deep learning, since forward and backward propagation is the foundation of CNN. By analyzing the key difference with whole-image classification, the proposed algorithms eliminate all the redundant computation in the forward and backward propagation in CNN for pixelwise classification. With guarantee on producing exactly the same results as patch-by-patch scanning, over $1,500$ times speedup has been achieved in our experiments, and the speedup will further increase with the sizes of images and patches. The proposed $d$-regularly sparse kernels can  covert convolution and pooling kernels with various strides into operations with 1-strides, which allows continuous memory access on GPU. Therefore, it has great flexibility to handle various CNNs with different designs and structures, and reaches high efficiency on GPU implementation.  

It opens the door to many high-impact applications and research in the future. It breaks the efficiency bottleneck of CNN based pixelwise classification,
and makes the realtime applications possible. It  has the potential to change  CNN training  fundamentally. Since at each training iteration the error map over all pixels in an image can be estimated quickly with fast forward propagation, it can be used to guide the selection of training patches by considering the spatial distribution and dynamic variation of errors. In contrast, training patches were completely randomly selected in  existing works. Moreover, an arbitrary subset of training patches can be selected from an image for the proposed fast backward propagation with constant computation complexity. Many interesting training strategies are expected to be developed based our work.  
It is not difficult to extend our algorithms to video analysis with 3D convolution and 3D pooling, where much more redundancy exists in cube-by-cube scanning and even  higher speedup is expected.



{\small
\bibliographystyle{ieee}
\bibliography{ref}
}

\end{document}